\documentclass[10pt,twocolumn,letterpaper]{article}

\usepackage{cvpr}              

\usepackage{graphicx}
\usepackage{booktabs, tabularx}
\usepackage{adjustbox}
\usepackage{array}

\usepackage{listings}
\usepackage{xcolor}

\definecolor{codegray}{RGB}{40,40,40}
\definecolor{commentgray}{RGB}{140,140,140}
\definecolor{keywordblue}{RGB}{0,90,170}
\definecolor{stringred}{RGB}{190,50,50}
\definecolor{cvprblue}{rgb}{0.21,0.49,0.74}

\usepackage[pagebackref,breaklinks,colorlinks,allcolors=cvprblue]{hyperref}

\usepackage{booktabs}
\usepackage{multirow}
\usepackage{xcolor}

\usepackage{algorithm}
\usepackage{algpseudocode}
\usepackage[table]{xcolor}  

\usepackage{multirow}
\usepackage{booktabs}
\usepackage{makecell}
\usepackage[table]{xcolor}
\definecolor{mygray}{gray}{0.9}
\definecolor{mygrayL}{gray}{0.95}

\usepackage{pifont}
\usepackage[dvipsnames]{xcolor}

\usepackage[absolute,overlay]{textpos}
\usepackage{dblfloatfix}

\definecolor{cvprblue}{rgb}{0.21,0.49,0.74}
\usepackage[pagebackref,breaklinks,colorlinks,allcolors=cvprblue]{hyperref}

\def\paperID{****} 
\def\confName{CVPR}
\def\confYear{2026}

\title{DynaTok: Temporally Adaptive and Positional Bias-Aware Token Compression for Video-LLMs}

\author{
  Minyoung Park \quad Taehun Kong \quad Sangjun Ahn\thanks{Corresponding author.} \\
  LG Electronics, Seoul, South Korea \\
  {\tt\small minyoung5.park@lge.com \quad th.kong@lge.com \quad sangjun.ahn@lge.com}
}

\begin{document}
\maketitle
\begin{abstract}
Recent advances in Video Large Language Models (Video-LLMs) have greatly expanded multimodal reasoning capabilities. However, the massive number of visual tokens extracted from long video sequences incurs prohibitive computational costs, limiting their deployment in real-world scenarios. Existing training-free token compression methods select tokens based on attention magnitude as a proxy for semantic importance, but often overlook positional bias and rely only on short-term temporal locality, leading to redundant spatio-temporal coverage and inefficient token usage.
We present DynaTok, a training-free, temporally adaptive and bias-aware token compression framework that allocates token budgets across both temporal and spatial dimensions. Through a lightweight exponential moving average (EMA) memory, the Temporal Budget Allocation (TBA) module dynamically assigns fewer tokens to redundant frames and more to novel frames, capturing long-term temporal variation. The Spatial Budget Allocation (SBA) module complements this by selecting spatially diverse and semantically important features using activation-based attention maps, while leveraging a spatial memory to reduce redundancy from previously selected regions and mitigate positional bias.
DynaTok integrates seamlessly with existing Video-LLMs such as LLaVA-OneVision and LLaVA-Video without retraining, and effectively preserves semantic coverage under aggressive compression. Experiments on four representative VideoQA benchmarks—MVBench, LongVideoBench, MLVU, and VideoMME—show that DynaTok retains over 95\% of baseline accuracy even with a 90\% token reduction, surpassing recent training-free approaches. These results demonstrate that DynaTok provides a principled foundation for efficient and robust video reasoning, paving the way toward real-time streaming video understanding with future Video-LLMs.
\end{abstract}
    
\label{sec:intro}

\begin{figure}[t]
    \centering
    \includegraphics[width=\linewidth, trim={300 135 280 120}, clip]{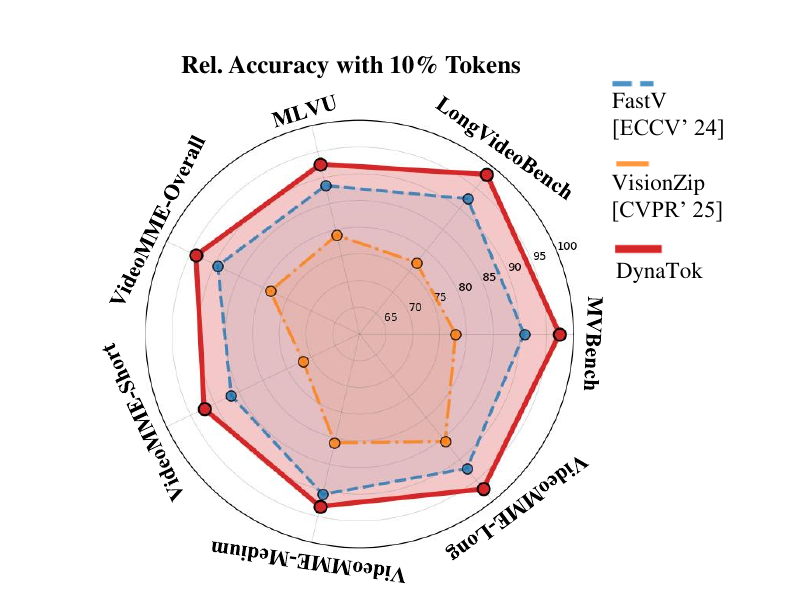}
    \caption{
Relative accuracy (\%) of token compression methods under a 10\% token budget on LLaVA-OneVision 7B.  
We report performance across MVBench~\citep{li2024mvbench}, LongVideoBench~\citep{wu2024longvideobench}, MLVU~\citep{zhou2024mlvu}, and VideoMME~\citep{fu2025video}.  
Relative accuracy is computed with respect to the base model's performance when using 100\% of input tokens.  
Compared to existing methods such as FastV~\citep{chen2024image} and VisionZip~\citep{yang2025visionzip}, which experience performance degradation under aggressive compression, DynaTok demonstrates consistently higher robustness and achieves the best relative accuracy across all benchmarks.
}
    \label{fig:first_figure}
\end{figure}

\begin{figure*}[t]  
  \centering
  \includegraphics[width=1\textwidth, trim={110 100 110 75}, clip]{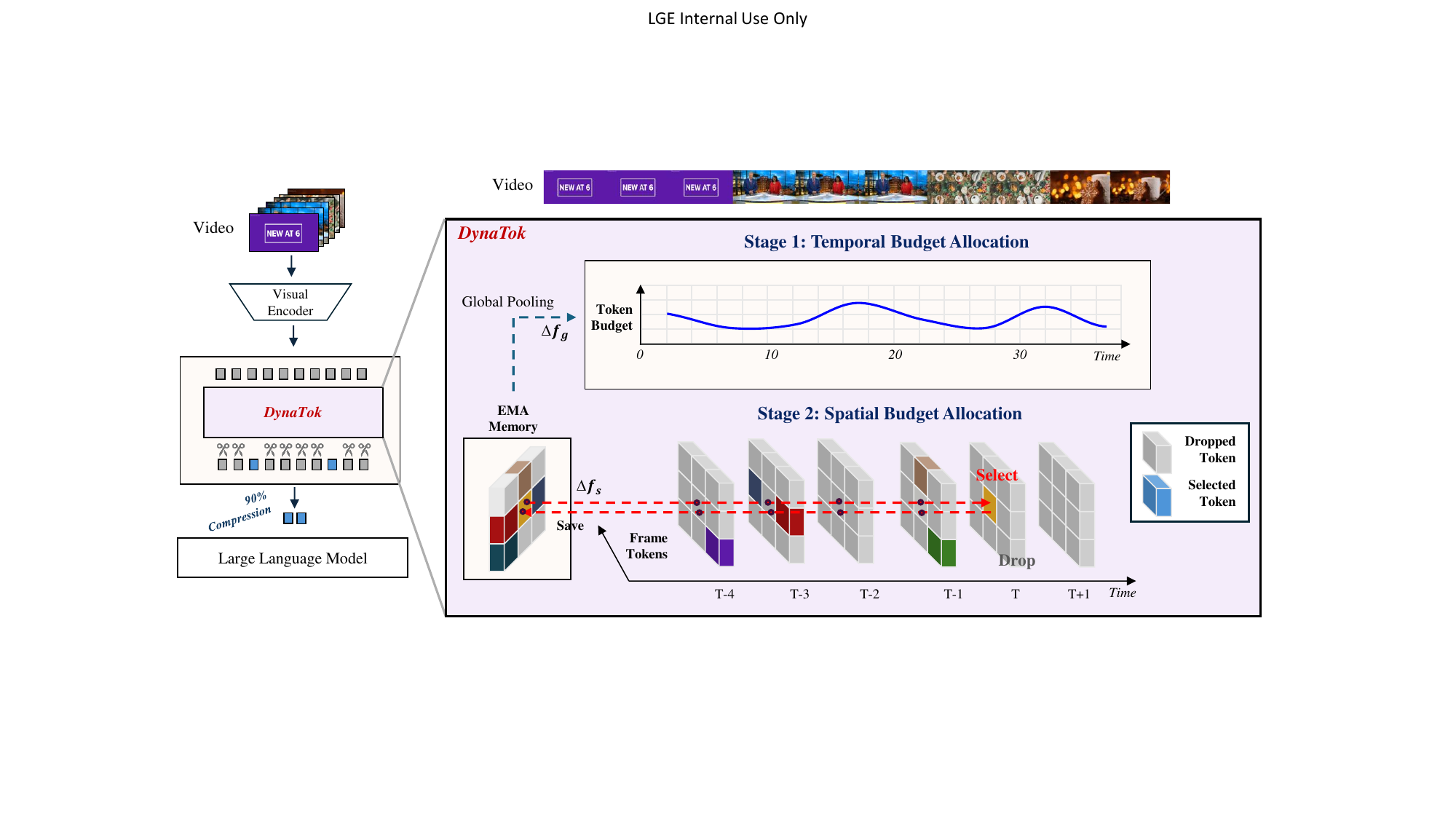}
\caption{
        Overview of the proposed DynaTok framework for training-free and efficient token compression in Video-LLMs. The framework operates in two stages: 
        (1) Temporal budget allocation (TBA) uses an exponential moving average (EMA)-based global temporal memory to estimate frame-wise novelty. It allocates fewer tokens to redundant frames and more to frames with dynamic or novel content based on global feature differences.
        (2) Spatial budget allocation (SBA) then performs intra-frame token selection using activation-based attention maps and a spatial memory that reduces the influence of previously selected redundant regions. This promotes the selection of semantically informative and spatially diverse tokens.
        Together, these stages enable DynaTok to achieve high compression ratios while preserving critical spatio-temporal information for robust video-language understanding.
}

  \label{fig:model_overview}
\end{figure*}

\section{Introduction}

Recent advances in video large language models (Video-LLMs) have significantly enhanced multimodal reasoning capabilities, enabling models to process long and complex video sequences~\citep{bai2023qwen, team2024gemini, wang2024qwen2, zhang2023video, li2024llava, lin2024video, ren2024timechat}.  
However, the large number of visual tokens extracted from video frames imposes severe computational costs~\citep{song2024moviechat, maaz2024video, zhang2025videollama, weng2024longvlm}.  
Because the attention complexity grows quadratically with the token count, efficient token compression is essential for achieving practical real-time or on-device Video-LLMs.
While FlashAttention~\citep{dao2022flashattention, dao2023flashattention} has alleviated the quadratic cost of attention by reducing it to linear complexity, the overhead of processing a large number of video input tokens—particularly during key-value (KV) cache pre-filling—remains a bottleneck. As a result, several recent works~\citep{chen2024image,yang2025visionzip,yao2025timechat,jeddi2025similarity} have explored training-free token compression methods that aim to reduce redundancy without requiring model retraining.

Recent approaches such as FastV~\cite{chen2024image}, DyCoke~\cite{tao2025dycoke}, and VisionZip~\cite{zhang2025videollama} leverage attention magnitude as a proxy for semantic importance, offering plug-and-play solutions for accelerating inference. However, these methods often overlook positional bias inherent to the vision encoders in Video-LLMs, which can significantly affect the quality of token selection.

Although attention scores are commonly interpreted as indicators of semantic relevance, prior studies~\citep{zou2025don, liu2025vrope, tian2025identifying, shen2025llava} have demonstrated that vision encoders in Video-LLMs inherit strong positional bias. This phenomenon arises from the joint training of visual and textual tokens using unified attention mechanisms in large language models. In the text modality, positional information is frequently aligned with semantic importance (e.g., key verbs or subjects appear in particular positions), and such position-based biases can be transferred to the visual modality during cross-modal training. As a result, certain spatial locations—such as image boundaries or bottom corners—tend to receive disproportionately high attention scores regardless of their actual semantic content. We also empirically confirm the presence of such positional bias in Video-LLMs (Fig.~\ref{fig:position_bias}).

Consequently, relying solely on attention magnitude for token pruning can be problematic—especially under aggressive compression—since attention-biased regions may dominate token retention, potentially leading to significant performance degradation. To address this, we propose a \textit{bias-aware token selection strategy} that not only considers semantic importance but also promotes spatio-temporal diversity in the selected tokens.

Another key challenge lies in handling temporal locality. In video-language modeling, redundancy across consecutive frames is often unevenly distributed. Recent approaches~\cite{shen2025fastvid, hyun2025multi} attempt to reduce tokens by exploiting short-term temporal similarity among adjacent frames. However, these methods typically operate within a limited temporal window, making it difficult to incorporate information from earlier frames beyond that range. This limitation becomes particularly problematic in streaming or real-time settings, where long-term context is critical.

To overcome this limitation, we propose a lightweight exponential moving average (EMA) memory module that incrementally aggregates spatio-temporal features from previously observed frames. This design enables our framework to estimate long-range frame novelty and adaptively allocate token budgets over time.

To address both positional and temporal inefficiencies, we introduce DynaTok, a temporally adaptive and bias-aware token compression framework that dynamically allocates token budgets across both the temporal and spatial dimensions, without requiring any retraining or architectural modification.

DynaTok operates in two complementary stages:

\begin{itemize}
    \item \textbf{Temporal Budget Allocation (TBA)}:  
    A lightweight temporal memory updated via an exponential moving averag estimates frame-wise novelty and dynamically allocates fewer tokens to redundant frames and more to frames with new or rapidly changing content.

    \item \textbf{Spatial Budget Allocation (SBA)}:  
    Within each frame, DynaTok leverages activation-based attention maps and a spatial memory to select tokens that are both semantically informative and spatially diverse, while reducing the influence of previously selected redundant regions.
\end{itemize}

This two-stage design enables flexible token allocation across time and space, allowing DynaTok to retain key semantics and long-term temporal information while significantly reducing computational cost. As shown in Fig.~\ref{fig:first_figure}, DynaTok achieves the highest relative accuracy under a 10\% token budget across all benchmarks, consistently outperforming training-free methods such as FastV~\citep{chen2024image} and VisionZip~\citep{yang2025visionzip}.

In summary, DynaTok provides a simple yet effective solution for training-free video token compression by explicitly addressing two fundamental sources of inefficiency—positional bias and temporal redundancy. Its temporally adaptive and bias-aware design enables efficient token selection while reducing redundancy without severely compromising semantic fidelity, offering a scalable and practical foundation for Video-LLM inference.

\section{Related Work}
\label{sec:rel_work}

\subsection{Token Efficiency for Video-LLMs}

Recent advances in Video-LLMs~\citep{zhang2023video, maaz2024video, li2023videochat, bai2023qwen, team2024gemini, wang2024qwen2, zhang2025videollama} have significantly expanded the capabilities of large language models for video understanding tasks, including video question answering, captioning, and temporal reasoning. To capture the rich spatio-temporal structure of video inputs, these models typically tokenize each frame into hundreds of patches, resulting in thousands to tens of thousands of visual tokens per video. This leads to substantial computational and memory overhead, often exceeding the context length limitations of LLMs.

Although methods like FlashAttention~\citep{dao2022flashattention,dao2023flashattention} improve the efficiency of attention computation by reducing its complexity from quadratic to near-linear, they do not mitigate the overhead associated with processing and caching a large number of visual tokens—particularly during key-value (KV) cache pre-filling. As a result, token quantity remains a critical bottleneck, especially in latency-sensitive or resource-constrained scenarios.

To address this challenge, recent works have proposed token compression strategies aimed at reducing visual redundancy and improving inference efficiency.

In parallel, there is growing interest in adapting Video-LLMs for real-time and streaming scenarios~\citep{chen2024videollm, ding2025streammind,li2025lion}, where low-latency inference and efficient context reuse become essential. In such settings, key-value caches must often be reused across turns, and any token compression strategy must be capable of reducing temporal redundancy while supporting long-form interaction. TimeChat-Online~\citep{yao2025timechat} further suggests that a large portion of visual tokens in streaming video are naturally redundant, reinforcing the importance of compression techniques that preserve semantic fidelity while reducing input volume over time.

In this work, we propose a training-free, model-agnostic token compression framework that significantly reduces the number of visual tokens while maintaining robust performance across multiple Video-LLM architectures and benchmarks. By leveraging a lightweight EMA memory to accumulate longer temporal context, our approach is naturally extendable to real-time streaming applications where efficient token management is crucial.

\subsection{Training-Free Compression with Spatio-Temporal Diversity}

Recent studies have explored training-free token compression strategies for Video-LLMs.
FastV~\citep{chen2024image} removes visual tokens with low attention weights using a plug-and-play approach.
SparseVLM~\citep{zhang2024sparsevlm} identifies a subset of text tokens that are highly related to the image and uses their cross-attention to score visual tokens.
DyCoke~\citep{tao2025dycoke} extends this line of work to video-language models, dynamically pruning redundant visual tokens during decoding based on attention scores across temporal frames.
However, these methods rely on internal attention matrices from the LLM backbone, which restricts their compatibility with FlashAttention~\citep{dao2022flashattention,dao2023flashattention}, as the required attention weights are not directly accessible due to fused computations.
Our method avoids this limitation by leveraging activation-based attention maps to score token importance. This enables a fully model-agnostic and FlashAttention-compatible token selection process that does not depend on internal attention matrices.
In contrast, VisionZip~\citep{yang2025visionzip} selects dominant visual tokens and merges semantically redundant ones using visual attention scores from the visual encoder, making it compatible with FlashAttention. While this design improves scalability, relying solely on attention magnitude has limitations—especially under aggressive compression. One key reason is the positional bias inherently present in the visual encoders of Video-LLMs.
Recent studies~\citep{zou2025don, liu2025vrope, tian2025identifying,xia2025video,shen2025llava} show that vision encoders in vision language models inherit strong positional bias, which is shifted from the text modality due to shared attention layers in joint training. This results in spatially biased attention patterns—e.g., consistently favoring certain image regions—regardless of semantic content. As a result, methods that depend only on attention scores may perform suboptimally under high compression. To address this, we propose a bias-aware token selection strategy that reduces the influence of previously selected redundant regions.
In addition to spatial bias, another key inefficiency arises from temporal redundancy. While several recent works~\citep{shen2025fastvid,hyun2025multi,tao2025dycoke} attempt to exploit temporal locality by comparing adjacent frames, their reliance on short-term context limits performance in long-horizon or streaming scenarios. These models typically apply redundancy reduction within a fixed temporal window, making it difficult to adaptively budget tokens across varying temporal dynamics.
To overcome this, we introduce a lightweight EMA memory that maintains spatio-temporal summaries of previously seen frames. This allows our method to detect novelty over longer time spans and allocate more tokens to novel or fast-changing frames—without requiring future frame access.
In summary, DynaTok combines a bias-aware spatial selection strategy with temporally adaptive token budgeting, yielding a model-agnostic compression framework that retains semantic and temporal fidelity under strict token budgets. Our approach is also scalable to streaming video understanding tasks.

\section{Method}

\begin{figure}[t]
    \centering
    \includegraphics[width=\linewidth, trim={340 200 340 200}, clip]{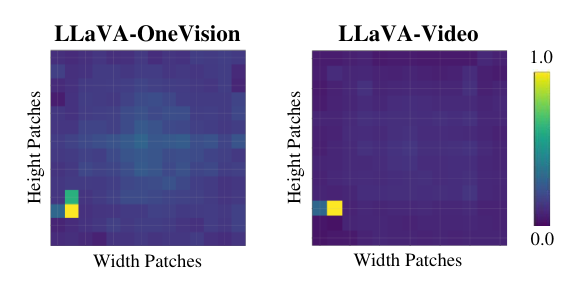}
    \caption{
    Visualization of accumulated activation-based attention maps on the VideoMME dataset. 
    While these maps reflect semantic importance, they also reveal strong positional bias, 
    where certain spatial regions consistently receive higher responses regardless of semantic content. 
    This indicates that attention magnitude alone is insufficient for token selection, 
    motivating the need to also preserve feature diversity in DynaTok.}
    \label{fig:position_bias}
\end{figure}

\subsection{Overview}

\begin{algorithm}[t]
\caption{DynaTok: Two-Stage Dynamic Token Compression}
\label{alg:dynatok}
\small
\begin{algorithmic}[1]
\Require Video tokens $X \in \mathbb{R}^{T \times N \times D}$, number of patches $P$, retention ratio $R$, weights $\alpha$, $\beta$
\Statex
\# \textbf{\textcolor{RoyalBlue}{1. Temporal Budget Allocation (TBA)}}
\State Initialize temporal memory $M_g \leftarrow \text{GlobalPool}(X_0)$
\For{$t = 1$ to $T$}
    \State global diff. $\Delta f_g \leftarrow \| \text{GlobalPool}(X_t) - M_g \|_2$
    \State Update $M_g \leftarrow (1 - \alpha) M_g + \alpha\,\text{GlobalPool}(X_t)$
\EndFor
\State Normalize $\{\Delta f_g\}$ to obtain frame-wise weights $w_t$
\State Assign token budget for each frame: $B_t = w_t \cdot T \cdot N \cdot R$
\Statex
\# \textbf{\textcolor{RoyalBlue}{2. Spatial Budget Allocation (SBA)}}
\State Initialize spatial memory $M_s \leftarrow \text{zeros\_like}(X_0)$
\For{each frame $t$}
    \State Divide $X_t$ into patches $\{p_i\}_{i=1}^P$
    \State Activation-based attention map: $A_t = \text{norm}\left( \sum_D |X_t| \right)$
    \State Patch importance: $s_i = \text{mean}(A_t[p_i])$
    \State Normalize $\{s_i\}$ to assign token counts $n_i$ s.t. $\sum_i n_i = B_t$
    \For{each token $x$ in $X_t$}
        \State Redundancy $\Delta f_s[x] = \text{cosine}(x, M_s[x])$
        \State Combined $\text{score}(x) = A_t[x] - \beta\,\Delta f_s[x]$
    \EndFor
    \For{each patch $p_i$}
        \State Select top-$n_i$ tokens within $p_i$ based on $\text{score}(x)$
    \EndFor
    \State $M_s[K_t] \leftarrow (1 - \alpha) M_s[K_t] + \alpha X_t[K_t]$ 
    \Statex \hspace{\algorithmicindent} \quad \textit{\small ($K_t$: selected token mask)}
\EndFor
\State \Return Compressed tokens $X' = X[K]$
\end{algorithmic}
\end{algorithm}

We propose DynaTok, a training-free two-stage framework that operates under a fixed token budget and efficiently allocates tokens across spatial and temporal dimensions to reduce redundancy in video inputs for Video-LLMs. 
As illustrated in Fig.~\ref{fig:model_overview}, DynaTok comprises two stages:
(1) Temporal Budget Allocation (TBA) and
(2) Spatial Budget Allocation (SBA).
In the first stage, DynaTok dynamically distributes the overall token budget across frames based on frame-wise novelty, estimated using a temporal memory updated via an exponential moving average.
In the second stage, it selects semantically informative yet spatially diverse tokens within each frame, guided by activation-based attention maps and a spatial memory that reduces the influence of previously selected redundant regions.
This two-stage design ensures effective token utilization under strict retention constraints, while preserving essential semantic and temporal cues—achieving a strong balance between compression efficiency and reasoning accuracy.

\subsection{Stage 1: Temporal Budget Allocation}
The first stage allocates token budgets along the temporal axis.
Since consecutive frames in a video often contain redundant visual information, DynaTok assigns fewer tokens to such redundant frames and redistributes the saved budget to frames containing novel or dynamic content.

Let the input video tokens be $X \in \mathbb{R}^{T \times N \times D}$, where $T$ is the number of frames, $N$ is the number of tokens per frame, and $D$ is the feature dimension.
We maintain a temporal memory $M_g$ that captures long-term temporal dynamics through an exponential moving average (EMA).
For each frame $t$, the global feature difference is computed as:
\begin{equation}
\Delta f_g = \left\| \text{GlobalPool}(X_t) - M_g \right\|_2,
\end{equation}
and the temporal memory is updated as:
\begin{equation}
M_g \leftarrow (1 - \alpha) M_g + \alpha \cdot \text{GlobalPool}(X_t).
\end{equation}
Here, $\alpha$ is a smoothing factor that controls the update rate. This EMA update emphasizes recent frame features while still retaining a cumulative summary of past observations.
The differences $\Delta f_g$ are normalized across frames to obtain frame-wise weights $w_t$, and the corresponding token budget for each frame is assigned as:
\begin{equation}
B_t = w_t \cdot T \cdot N \cdot R,
\end{equation}
where $R$ denotes the global retention ratio.
Consequently, frames exhibiting higher temporal variation (larger $\Delta f_g$) receive more tokens, promoting temporal diversity and improving compression efficiency.

\subsection{Stage 2: Spatial Budget Allocation}

After distributing the token budget temporally, DynaTok performs spatial selection within each frame to determine which tokens to retain. This stage focuses on identifying semantically important yet spatially diverse regions while avoiding redundant coverage.

To estimate token importance, we compute an activation-based attention map:
\begin{equation}
A_t = \text{norm}\left(\sum_{D} |X_t|\right),
\end{equation}
which serves as a model-agnostic measure of semantic saliency. Unlike prior approaches~\citep{chen2024image,zou2025don} that rely on internal attention weights or CLS tokens, this formulation is compatible with modern encoders (e.g., SigLIP~\citep{zhai2023sigmoid}) and FlashAttention implementations.

However, relying solely on semantic saliency can lead to positional bias, as certain spatial regions tend to receive consistently high activations due to cross-modal alignment during pretraining~\cite{zou2025don}. As shown in Fig.~\ref{fig:position_bias}, accumulated activation-based attention maps illustrate this phenomenon: while attention magnitude correlates with semantic importance, it also exhibits strong positional bias. This motivates the need for a diversity-aware selection strategy.

To ensure spatial diversity, we divide each frame $X_t \in \mathbb{R}^{N \times D}$ into $P = N / k$ non-overlapping patches $\{p_i\}_{i=1}^P$, each containing $k$ tokens. The semantic importance of each patch is computed as:
\begin{equation}
s_i = \text{mean}(A_t[p_i]),
\end{equation}
and the token budget $n_i$ for patch $p_i$ is assigned by normalizing importance across patches:
\begin{equation}
n_i = \frac{s_i}{\sum_j s_j} \times B_t, \quad \text{with } \sum_i n_i = B_t.
\end{equation}

To reduce the influence of previously selected redundant tokens, we introduce a spatial memory $M_s$ that stores tokens retained from earlier frames. Redundancy is penalized via cosine similarity with the spatial memory:
\begin{equation}
\Delta f_s[x] = \text{cosine}(x, M_s[x]),
\end{equation}
where $x$ denotes a token in $X_t$.

A combined score is then computed for each token:
\begin{equation}
\text{score}(x) = A_t[x] - \beta \Delta f_s[x],
\end{equation}
where $\beta$ controls the trade-off between semantic saliency and redundancy. From each patch $p_i$, we select the top-$n_i$ tokens based on this score.

Finally, selected tokens are used to update the spatial memory:
\begin{equation}
M_s[K_t] \leftarrow (1 - \alpha) M_s[K_t] + \alpha X_t[K_t],
\end{equation}
where $K_t$ is the binary mask indicating the selected tokens in frame $t$. This memory-guided selection promotes spatial diversity across frames while suppressing repeated focus on previously observed regions.

Together with the Temporal Budget Allocation stage, this spatial selection mechanism enables DynaTok to efficiently compress tokens while preserving semantic coverage and temporal reasoning performance.

\section{Experiments}

\begin{table*}[t]
\centering
\caption{Comparison of video token compression methods on LLaVA-OneVision 7B~\cite{li2024llava}. 
All methods are evaluated across multiple benchmarks, including MVBench, LongVideoBench, MLVU, and VideoMME.
The retention ratio $R$ denotes the proportion of visual tokens preserved before compression.
DynaTok maintains competitive performance even under extremely low token budgets,
showing strong robustness against aggressive compression while achieving the best overall accuracy across all settings.
}
\resizebox{\linewidth}{!}
{
\begin{tabular}{c|c|ccccccc|cc}
\toprule
 \multirow{2}{*}{Method} & \multirow{2}{*}{\makecell{Retention\\Ratio $R$}} & \multirow{2}{*}{MVBench} & \multirow{2}{*}{\makecell{LongVideo\\Bench}} & \multirow{2}{*}{MLVU} & \multicolumn{4}{c}{VideoMME}                              & \multicolumn{2}{|c}{Avg. Acc.} \\ \cline{6-9}
                                                       &                                 &                          &                                 &                       & Overall      & Short       & Medium       & Long          & Score       & \%         \\
\rowcolor{mygray} Duration                         &                                & 16s                      & 1$\sim$60min                    & 3$\sim$120min         & 1$\sim$60min & 1$\sim$3min & 3$\sim$30min & 30$\sim$60min &             &            \\ \midrule
\cellcolor{mygrayL}LLaVA-OV 7B                 & \cellcolor{mygrayL}100\%                         & \cellcolor{mygrayL}56.9                     & \cellcolor{mygrayL}56.4                            & \cellcolor{mygrayL}65.2                  & \cellcolor{mygrayL}58.6         & \cellcolor{mygrayL}70.3        & \cellcolor{mygrayL}56.6         & \cellcolor{mygrayL}48.8          & \cellcolor{mygrayL}59.3        & \cellcolor{mygrayL}100        \\ \hline
                                    DyCoke~\cite{tao2025dycoke}$_{\text{CVPR'25}}$ & 32.5\%                                & 56.3                     & 56.6                            & 62.1                  & 57.1         & 68.1        & 56.7         & 46.7          & 58.0        & 97.8        \\ \hline
                                  FastV~\cite{chen2024image}$_{\text{ECCV'24}}$                   &  100\%/25\%                            & 54.7                     & 55.5                            & 61.5                  & 56.2         & 68.0        & 54.6         & 46.0          & 57.0        & 96.1       \\
                                  VisionZip~\cite{yang2025visionzip}$_{\text{CVPR'25}}$               & 25\%                             & 53.7                     & 51.2                            & 58.5                  & 54.1         & 61.6        & 53.4         & 47.2          & 54.4        & 91.7       \\
                                  
                                  DyCoke~\cite{tao2025dycoke}$_{\text{CVPR'25}}$ & 25\%                             & 49.5                     & 48.1                            & 55.8                  & 51.0         & 61.1        & 48.6         & 43.2          & 51.1        & 86.2        \\
                                \textbf{DynaTok}             & 25\%                               & \textbf{57.0}                     & \textbf{56.3}                           & \textbf{62.5}                  &\textbf{57.8}	&\textbf{69.3}	&\textbf{55.1}	&\textbf{49.0}  &   \textbf{58.4}    &   \textbf{98.6}            \\ 
                                  
                                  \hline
                                  FastV~\cite{chen2024image}$_{\text{ECCV'24}}$                   & 100\%/20\%          & 54.1                     & \textbf{56.6}                            & 61.2                  & 56.2         & 66.8        & 54.6         & 47.2          & 57.0        & 96.1       \\
                                  VisionZip~\cite{yang2025visionzip}$_{\text{CVPR'25}}$               & 19.9\%                              & 53.0                     & 50.0                            & 57.1                  & 53.0         & 60.8        & 51.0         & 47.1          & 53.3        & 90.0       \\
                                  
                                  \textbf{DynaTok}             & 19.9\%                               & \textbf{56.7}                  & 55.7                         & \textbf{62.1}                   & \textbf{57.3}         & \textbf{68.7}       &\textbf{55.4}         & \textbf{47.8}          &     \textbf{58.0}     &   \textbf{97.9}    \\

                                  \hline 
                                  FastV~\cite{chen2024image}$_{\text{ECCV'24}}$                   & 100\%/15\%           & 53.2                     & 54.9                            & 59.8                  & 54.7         & 65.1        & 53.4         & 45.7          & 55.7        & 93.9       \\
                                  VisionZip~\cite{yang2025visionzip}$_{\text{CVPR'25}}$               & 14.8\%                               & 50.3                     & 46.9                            & 54.4                  & 49.5         & 55.8        & 49.3         & 43.3          & 50.3        & 84.8       \\
                                  
                                  \textbf{DynaTok}             & 14.8\%                       & \textbf{56.6}                     & \textbf{55.4}                           & \textbf{62.1}                  & \textbf{55.7}         & \textbf{67.0}        & \textbf{53.8}       & \textbf{46.4}          & \textbf{57.5}        & \textbf{97.0}      \\ 
                                  \hline
                                  FastV~\cite{chen2024image}$_{\text{ECCV'24}}$                   & 100\%/10\%           & 51.7                     & 52.1                            & 57.7                  & 52.4         & 60.9        & 51.4         & 45.0          & 53.5        & 90.2       \\
                                  VisionZip~\cite{yang2025visionzip}$_{\text{CVPR'25}}$               & 9.7\%                              & 44.4                     & 43.5                            & 51.5                  & 46.0         & 50.4        & 45.8         & 41.8          & 46.4        & 78.3       \\

                                  \textbf{DynaTok}            & 9.7\%                                & \textbf{55.4}  &  \textbf{55.4}& \textbf{60.4}& \textbf{55.0}& \textbf{64.8}& \textbf{52.7}& \textbf{47.4}         &  \textbf{56.6}        &  \textbf{95.5}     \\ 
                                  \bottomrule
\end{tabular}
}
\label{tab:sota_ov}
\vspace{-8pt}
\end{table*}
\begin{table*}[t]
\centering
\caption{Comparison of video token compression methods on LLaVA-Video 7B~\cite{zhang2024video}.
All methods are evaluated on MVBench, LongVideoBench, MLVU, and VideoMME benchmarks.
The retention ratio $R$ denotes the proportion of visual tokens preserved before compression.
DynaTok consistently maintains stable and high accuracy across all benchmarks and token budgets,
demonstrating robustness and model-agnostic generalization on the LLaVA-Video backbone.}
\resizebox{\linewidth}{!}
{
\begin{tabular}{c|cc|cccccc|cc}
\toprule
 \multirow{2}{*}{Method} & \multirow{2}{*}{\makecell{Retention\\Ratio $R$}} & \multirow{2}{*}{\makecell{\# Newline \\ Tokens $M$}} & \multirow{2}{*}{MVBench} & \multirow{2}{*}{\makecell{LongVideo\\Bench}} & \multirow{2}{*}{MLVU} & \multicolumn{3}{c}{VideoMME}                             & \multicolumn{2}{|c}{Avg. Acc.} \\ \cline{7-9}
                                                          &                                &                          &                                 &                       &             & Overall & Short & Long                  & Score       & \%         \\ 
\midrule
 \cellcolor{mygrayL}LLaVA-Video 7B                 & \cellcolor{mygrayL}100\% & \cellcolor{mygrayL}832            & \cellcolor{mygrayL}60.4                     & \cellcolor{mygrayL}59.6                            & \cellcolor{mygrayL}70.3                  & \cellcolor{mygrayL}64.1         & \cellcolor{mygrayL}76.9          & \cellcolor{mygrayL}53.4          & \cellcolor{mygrayL}63.6        & \cellcolor{mygrayL}100        \\ \hline 
                                  DyCoke~\cite{tao2025dycoke}$_{\text{CVPR'25}}$                  & 32.1\% & 256 & 59.3                     & 57.9                            & 65.7                  & 61.6         & 74.6          & 51.1          & 61.1        & 96.1       \\ \hline 
                                  
                                  FastV~\cite{chen2024image}$_{\text{ECCV'24}}$                   & 100\%/25\%      & 832/568.7 & 58.0                     & \textbf{58.3}                            & 63.9                  & 61.0         & 71.3               & 51.0          & 60.3        & 94.8       \\
                                  VisionZip~\cite{yang2025visionzip}$_{\text{CVPR'25}}$               & 24.9\%   & 64                            & 56.4                     & 54.1                            & 62.1                  & 58.6         & 66.3              & 51.2          & 57.8        & 90.9       \\
 DyCoke~\cite{tao2025dycoke}$_{\text{CVPR'25}}$                  & 25\% & 208           & 50.8                     & 53.0                            & 56.9                  & 56.1         & 65.8              & 48.9          & 54.2        & 85.2       \\
                                  DynaTok & 24.9\% & 832 & \textbf{58.6}&57.0&\textbf{64.6}&\textbf{62.1}&\textbf{74.1}&\textbf{51.6}&\textbf{60.6}&\textbf{95.4} \\ 
                                  \hline  FastV~\cite{chen2024image}$_{\text{ECCV'24}}$ & 100\%/10\% & 832/311.8 &	55.8	&	\textbf{55.4}	&	58.9	&	57.9	&	67.6	&	48.6	&	57.0	&	89.6  \\
                                  VisionZip~\cite{yang2025visionzip}$_{\text{CVPR'25}}$ & 9.5\% & 64 & 	46.3	&	46.6	&	52.2	&	49.5	&	54.2	&	44.3	&	48.7	&	76.6
  \\                                  DynaTok & 9.5\% & 832 &\textbf{55.9}&55.2&\textbf{63.2}&\textbf{59.0}&\textbf{70.6}&\textbf{49.7}&\textbf{58.3} &\textbf{91.8}\\
                                  \bottomrule
\end{tabular}
}
\label{tab:sota_vid}
\end{table*}
\begin{table}[t]
\centering
\small
\setlength{\tabcolsep}{3pt}
\caption{
Ablation study on the effectiveness of each stage in DynaTok.  
When both TBA and SBA are disabled (\ie, \ding{55} \ding{55}), the model performs frame-wise uniform token compression 
based solely on semantic importance.  
Enabling either stage improves performance, and combining both (\ding{51} \ding{51}) yields the best results, 
demonstrating that temporal and spatial budget allocation are complementary for efficient video token compression.
}

\begin{tabular}{ccccccc}
\toprule
\textbf{$R$ (\%)} & \textbf{TBA} & \textbf{SBA} & \textbf{VideoMME} & \textbf{Short} & \textbf{Medium} & \textbf{Long} \\
\midrule
25 & \ding{55} & \ding{55} & 56.9 & 68.4 & 55.3 & 46.9 \\
25 & \ding{51} & \ding{55} & 57.2 & 68.3 & 56.1 & 47.2 \\
25 & \ding{51} & \ding{51} & \textbf{57.8} & 69.3 & 55.1 & 49.0 \\
\midrule
9.7 &  \ding{55} & \ding{55} & 54.8 & 63.8 & 53.6 & 47.1 \\
9.7 &  \ding{51} & \ding{55} & 54.5 & 63.8 & 53.6 & 46.2 \\
9.7 & \ding{51} & \ding{51} & \textbf{55.0} & 64.8 & 52.7 & 47.4 \\
\bottomrule
\end{tabular}
\label{tab:ablation_sba_tba_stba}
\end{table}

\subsection{Experimental Setup}
We evaluate DynaTok on two recent Video-LLMs: 
LLaVA-OneVision 7B~\cite{li2024llava} and LLaVA-Video 7B~\cite{zhang2025videollama}.  
We conduct experiments on four representative Video Question Answering benchmarks:  
MVBench~\cite{li2024mvbench}, LongVideoBench~\cite{wu2024longvideobench},  
MLVU~\cite{zhou2024mlvu} and VideoMME~\cite{fu2025video}.  
Each model is evaluated under multiple token retention ratios ($R$)  
to analyze the trade-off between performance and computational efficiency, using the LMMS-Eval~\citep{zhang2025lmms} framework for standardized and reproducible evaluation.

\textbf{Baselines.}
We compare DynaTok against recent training-free token compression methods:  
FastV~\cite{chen2024image}, VisionZip~\cite{yang2025visionzip}, and DyCoke~\cite{tao2025dycoke}.

\textbf{Metrics.}  
We report the absolute accuracy on each benchmark, as well as the relative performance (\%) compared to the base Video-LLM using the full 100\% token budget.

\subsection{Results on LLaVA-OneVision 7B}

Tab.~\ref{tab:sota_ov} compares training-free video token compression methods under various token budgets, using LLaVA-OneVision 7B~\citep{li2024llava} as the base model.

DynaTok achieves consistently strong performance across all compression levels. Notably, even under an extremely low retention ratio of $R=10\%$, DynaTok maintains a high average accuracy of 95.5\%, outperforming all other methods under equivalent or even higher budgets. In the medium compression regime ($R=25\%$), it further boosts accuracy to 98.6\%, again ranking highest in overall accuracy. 

Compared to FastV~\citep{chen2024image} and VisionZip~\citep{yang2025visionzip}, DynaTok demonstrates superior robustness under aggressive compression. While FastV suffers notable degradation below 25\% retention, and VisionZip underperforms in long-duration scenarios (e.g., LongVideoBench and Long VideoMME), DynaTok retains both short- and long-term video reasoning capability with minimal accuracy loss.

These results highlight DynaTok’s ability to preserve semantic fidelity and temporal coherence under extreme token budget constraints, making it a practical and scalable solution for real-world, resource-constrained video understanding systems.

\subsection{Results on LLaVA-Video 7B}

Tab.~\ref{tab:sota_vid} presents the evaluation results of token compression methods on LLaVA-Video 7B~\citep{zhang2024video} across multiple video-language benchmarks. While the vanilla model achieves an upper-bound accuracy of 100\% using all tokens, the focus of this comparison is on assessing how well each method balances compression efficiency with performance robustness across varying tasks and video durations.

DynaTok demonstrates strong and consistent performance across all benchmarks and compression levels. Even under an aggressive 9.5\% retention ratio, it achieves the highest average accuracy (91.8\%), significantly outperforming other methods such as FastV~\citep{chen2024image} (89.6\%) and VisionZip~\citep{yang2025visionzip} (76.6\%) under the same token budget.

At moderate compression levels (e.g., 25\%), DynaTok maintains overall strong accuracy across benchmarks. While its performance on LongVideoBench is slightly lower than FastV, it still demonstrates robust and stable results. In particular, DynaTok achieves higher scores on MLVU (64.6) and VideoMME-Long (51.6), both of which require long-term temporal understanding, underscoring its effectiveness on extended video content.

\begin{figure}[t]
    \centering
    \hfill
    \includegraphics[width=\linewidth, trim={315 150 285 130}, clip]{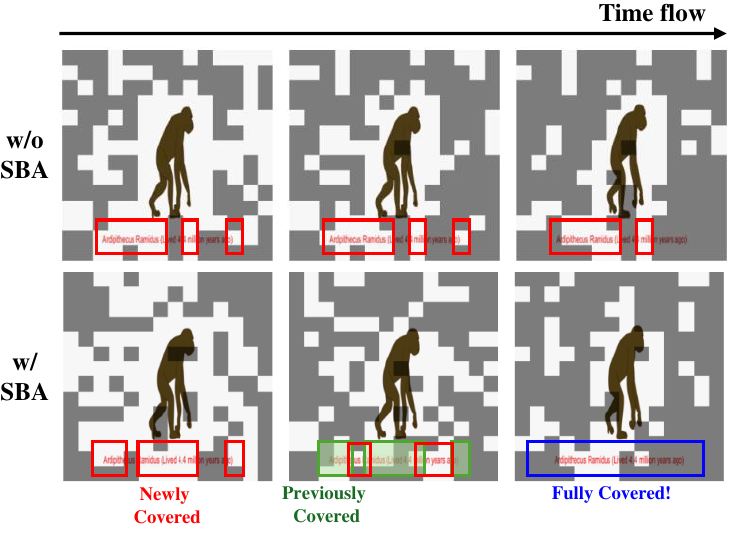}
    \caption{
    Visualization of token selection with and without the proposed Spatial Budget Allocation (SBA).
    Without SBA, token selection retains only the most semantically important regions at each frame,
    causing redundant focus on the same areas over time.
    With SBA (which reduces the influence of previously selected redundant regions), DynaTok prioritizes tokens that have not been previously covered, allowing the model to capture more diverse information across frames.}

    \label{fig:stba_ablation}
\end{figure}

\subsection{Ablation Study} 

Tab.~\ref{tab:ablation_sba_tba_stba} presents the quantitative results of our ablation study. When both TBA and SBA are disabled, the model performs uniform token compression across frames based solely on attention scores, which serve as a proxy for semantic importance but overlook positional bias.

Under a moderate compression setting ($R=25\%$), enabling either TBA or SBA yields noticeable improvements across overall score, validating the individual effectiveness of temporal and spatial guidance in token allocation. In contrast, under an aggressive compression setting ($R=9.7\%$), combining both TBA and SBA achieves the best performance, demonstrating their complementary roles in maintaining performance under extreme token constraints.

\begin{figure*}[t]
    \centering
    \includegraphics[width=\textwidth, trim={130 90 130 50}, clip]{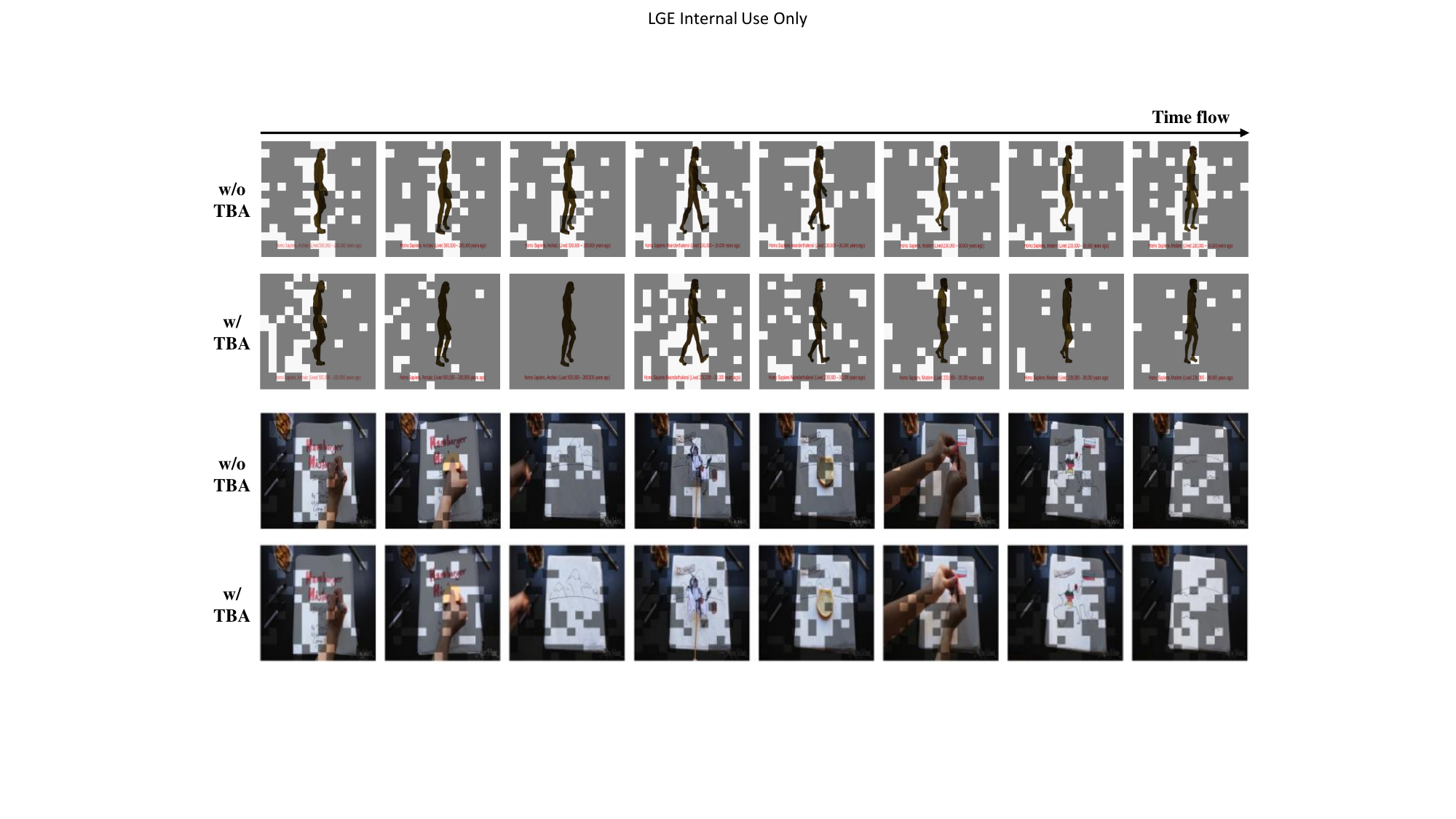}
    \caption{
    Visualization of the effect of Temporal Budget Allocation (TBA).
    Without TBA (frame-wise uniform token compression), an equal number of tokens is assigned to every frame, 
    regardless of temporal redundancy.  
    With TBA, DynaTok can dynamically allocate token budgets across frames — assigning fewer tokens to redundant frames 
    and more tokens to frames containing novel or dynamic content — thereby utilizing the token budget more efficiently 
    and focusing computation on temporally informative regions.}
    \label{fig:qualitative}
\end{figure*}

\subsection{Qualitative Analysis of Temporal and Spatial Coverage} 

Fig.~\ref{fig:qualitative} and Fig.~\ref{fig:stba_ablation} provide a qualitative analysis of how the proposed Temporal Budget Allocation (TBA) and Spatial Budget Allocation (SBA) influence token selection over time. 
These visualizations illustrate how DynaTok adaptively balances token allocation across frames and regions, achieving both temporal efficiency and spatial diversity.

Fig.~\ref{fig:qualitative} illustrates the impact of TBA.
In the absence of TBA (i.e., with uniform token allocation across frames), each frame receives an equal number of tokens,
regardless of its temporal redundancy, resulting in inefficient token utilization.
With TBA, DynaTok dynamically adjusts the number of retained tokens according to frame-wise temporal variation— 
allocating fewer tokens to redundant frames and more to those containing novel or motion-rich content. 
This adaptive allocation enables the model to concentrate computation on temporally informative regions, 
resulting in more efficient token utilization and improved capture of temporal dynamics across video sequences.

Fig.~\ref{fig:stba_ablation} visualizes the effect of the proposed SBA. 
Without SBA, token selection is dominated by the most semantically salient regions in each frame, 
leading to repetitive focus on similar spatial areas across time and limited scene coverage. 
With SBA, which incorporates a redundancy penalty through spatial memory, 
previously selected regions are deprioritized while new and complementary areas are encouraged. 
This promotes spatial diversity and allows DynaTok to capture a broader range of visual cues across frames, 
resulting in richer and more comprehensive scene understanding.

Together, these qualitative observations are consistent with the quantitative results, 
demonstrating that jointly modeling temporal and spatial redundancy is key to achieving efficient yet informative video token compression. 
By dynamically allocating tokens across both time and space, DynaTok provides temporally adaptive and bias-aware token selection, ensuring rich spatio-temporal coverage while maintaining strong reasoning performance even under tight token constraints.

\section{Conclusion}
We introduced DynaTok, a training-free and model-agnostic token compression framework for Video-LLMs operating under constrained token budgets. Unlike prior methods that rely solely on attention magnitude—often suffering from positional bias and being limited to short temporal windows—DynaTok employs a two-stage compression strategy: Temporal Budget Allocation and Spatial Budget Allocation. By integrating lightweight memory modules that track both global and local token distributions over time, our method achieves temporally adaptive and bias-aware token selection, allocating more tokens to novel and informative regions while reducing redundancy across both space and time.
DynaTok consistently outperforms recent training-free baselines across different Video-LLM architectures and benchmarks, including LLaVA-OneVision and LLaVA-Video, demonstrating strong generalization across diverse video understanding tasks. It remains robust even under extreme compression settings (e.g., 9.7\% token retention), maintaining high reasoning accuracy without requiring any retraining or architectural modification. This highlights the practical usability of DynaTok as a plug-and-play component in existing Video-LLM pipelines.
These results highlight the effectiveness of our approach and its potential for real-time applications under resource constraints.


\maketitlesupplementary

\renewcommand{\thesection}{S.\arabic{section}}
\renewcommand{\thesubsection}{S.\arabic{section}.\arabic{subsection}}
\renewcommand{\thefigure}{S\arabic{figure}}
\renewcommand{\thetable}{S\arabic{table}}

\begin{figure*}[!t]
\centering
\includegraphics[width=\linewidth, trim=220 200 220 200, clip]{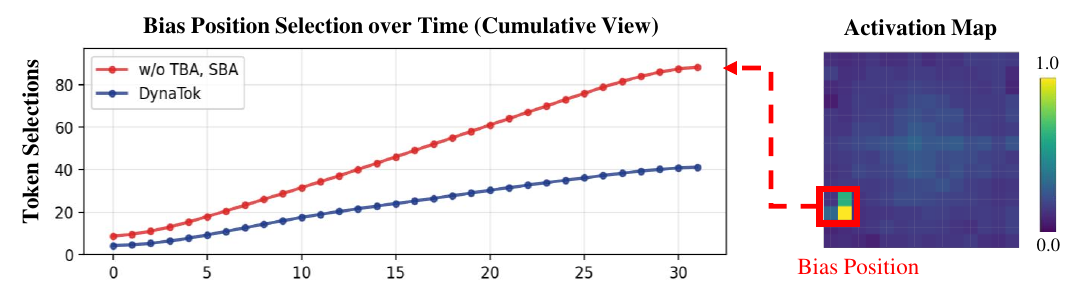}
\caption{Positional bias analysis on LLaVA-OneVision.
(Left) Cumulative token-selection curves at the known bias position across 32 frames. Without TBA/SBA (red), the model repeatedly selects the biased location, resulting in a steadily increasing selection count. With DynaTok (blue), the number of selections at the same position grows much more slowly, indicating reduced reliance on the biased region over time.
(Right) Example activation map with the high-bias region highlighted (red box), consistent with the positional bias observed in Fig. 3 of the main paper.}
\label{fig:supple_bias_cdf}
\end{figure*}

This supplementary document provides additional analysis and results supporting our main submission.
Unless otherwise stated, all experiments are conducted on LLaVA-OneVision 7B.

\section{Hyperparameters}
Table~\ref{tab:hyperparams} summarizes the hyperparameters used for integrating
DynaTok into Video-LLM architecture. The EMA decay factor $\alpha$ controls the
temporal smoothing in the memory update, while $\beta$ regulates the spatial
redundancy penalty in the spatial budget allocation module.

\begin{table}[h]
\centering
\begin{tabular}{l|c}
\hline
Parameter & Value \\
\hline
$\alpha$ & 0.9 \\
$\beta$  & 0.1 \\
Patch size $k$ & 14 tokens \\
Input frames & 32 \\
Hardware & 1$\times$H100\,80GB \\
\hline
\end{tabular}
\caption{
Hyperparameters used for DynaTok integration in Video-LLM. 
The parameters $\alpha$ and $\beta$ correspond to the temporal EMA decay and 
spatial redundancy penalty, respectively.
}
\label{tab:hyperparams}
\end{table}

\section{TBA Sensitivity to EMA Decay}

We sweep $\alpha \in \{0.70, 0.90, 0.95\}$ across four Video-LLM benchmarks at a 25\% token
retention ratio, as shown in Table~\ref{tab:alpha_sweep_multi}.

\begin{table}[h]
\centering
\begin{tabular}{c|c|c|c|c}
\hline
$\alpha$ & MVBench & LVB & MLVU & VideoMME \\
\hline
0.70 & 57.4 & 57.0 & 62.2 & 58.1 \\
0.90 & 56.7 & 55.7 & 62.1 & 57.3 \\
0.95 & 57.2 & 57.0 & 62.0 & 58.0 \\
\hline
\end{tabular}
\caption{TBA sensitivity to EMA decay $\alpha$ at 25\% retention.}
\label{tab:alpha_sweep_multi}
\end{table}

Lower $\alpha$ values preserve longer-term temporal history, whereas larger $\alpha$ values place
more emphasis on short-term frame differences. Across all benchmarks, performance varies within
1.3\%, indicating that DynaTok does not require fine-grained tuning of $\alpha$ and remains robust
over a wide temporal smoothing range.

\section{Latency and TTFT Efficiency}

We report end-to-end latency changes (including prefill and TTFT) as a function
of the retention ratio. At 100\% retention, the average latency is 1.94\,s on
our hardware. Table~\ref{tab:latency_reduction} shows the absolute reduction
in milliseconds when applying DynaTok at lower retention ratios, measured
relative to this 100\% baseline.

\begin{table}[h]
\centering
\begin{tabular}{c|c}
\hline
Retention & Latency Savings (ms) $\uparrow$ \\
\hline
25\% & 110 \\
10\% & 360 \\
\hline
\end{tabular}
\caption{
End-to-end latency reduction relative to the
100\% retention baseline (1.94\,s). Reported values indicate the absolute amount of latency saved.
}
\label{tab:latency_reduction}
\end{table}

\section{Visualizing Positional Bias Mitigation}
To validate that DynaTok mitigates the positional bias inherent in the vision encoder of Video-LLMs, we measure how often the model selects a known bias position across a 32-frame sequence. As shown in Fig.~\ref{fig:supple_bias_cdf} (left), the baseline model (w/o TBA, SBA) repeatedly selects the biased spatial location, causing the cumulative selection curve to rise almost linearly.

In contrast, Fig.~\ref{fig:supple_bias_cdf} (left) also shows that DynaTok substantially suppresses selections at the biased position. By adaptively redistributing the token budget across both time and space, DynaTok de-emphasizes tokens from biased regions and allocates more capacity to informative, non-biased areas. This results in a noticeably flatter cumulative curve.

Fig.~\ref{fig:supple_bias_cdf} (right) presents the activation map used to identify the positional bias region. The highlighted area corresponds to a consistently high-activation hotspot, matching the positional bias pattern observed previously in Fig.~3 of the main paper.

Overall, Fig.~\ref{fig:supple_bias_cdf} demonstrates that DynaTok effectively reduces positional bias over time. Both temporal budget allocation and redundancy-aware spatial budget allocation jointly prevent repeated selection of biased regions, even under the same global token budget.

\section{Full 32-Frame Visualizations}

\begin{figure*}[t]
\centering
\includegraphics[width=\linewidth]{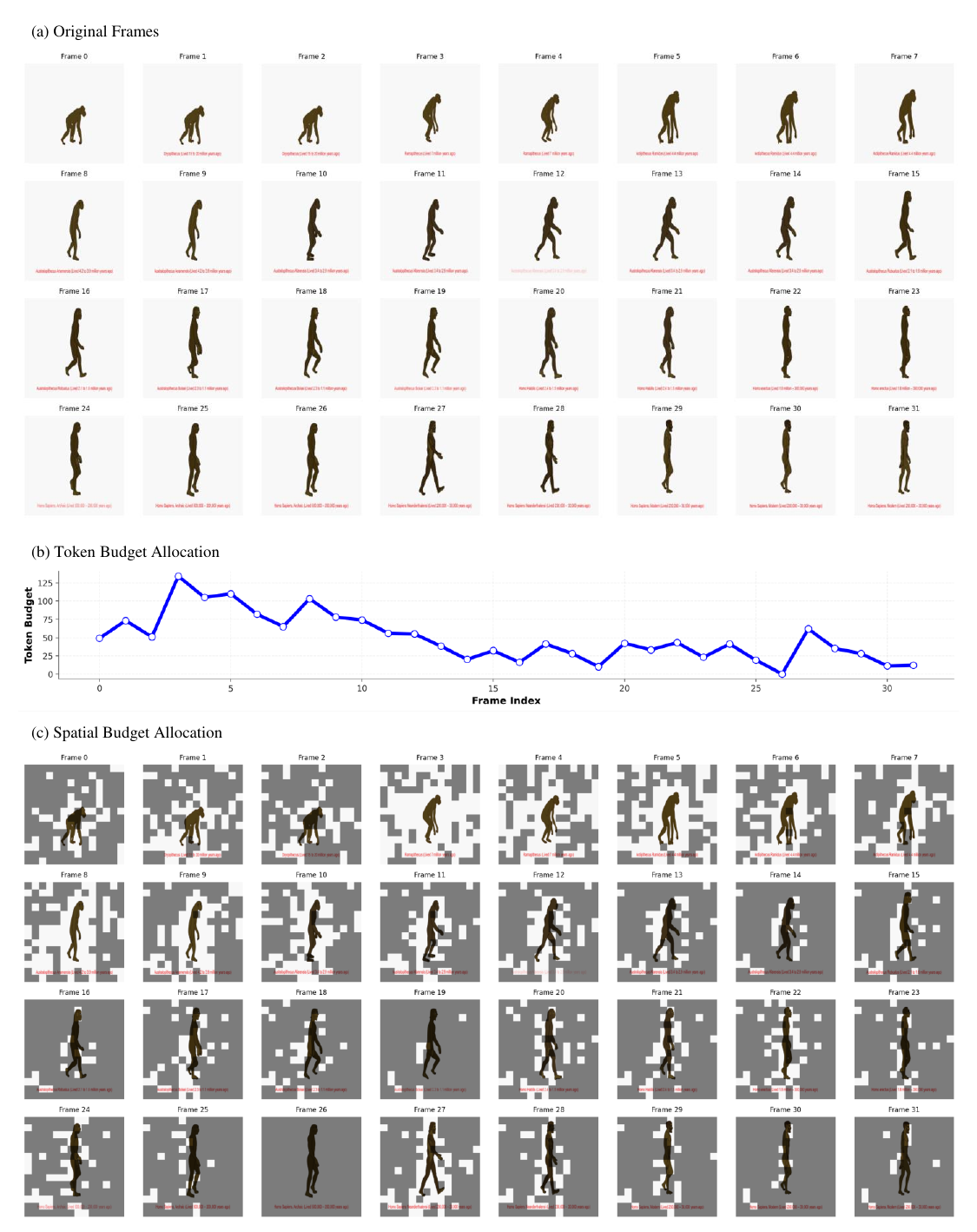}
\caption{Full 32-frame token compression visualization corresponding to the example in Fig. 5.}
\label{fig:supple_figures_human}
\end{figure*}

\begin{figure*}[t]
\centering
\includegraphics[width=\linewidth]{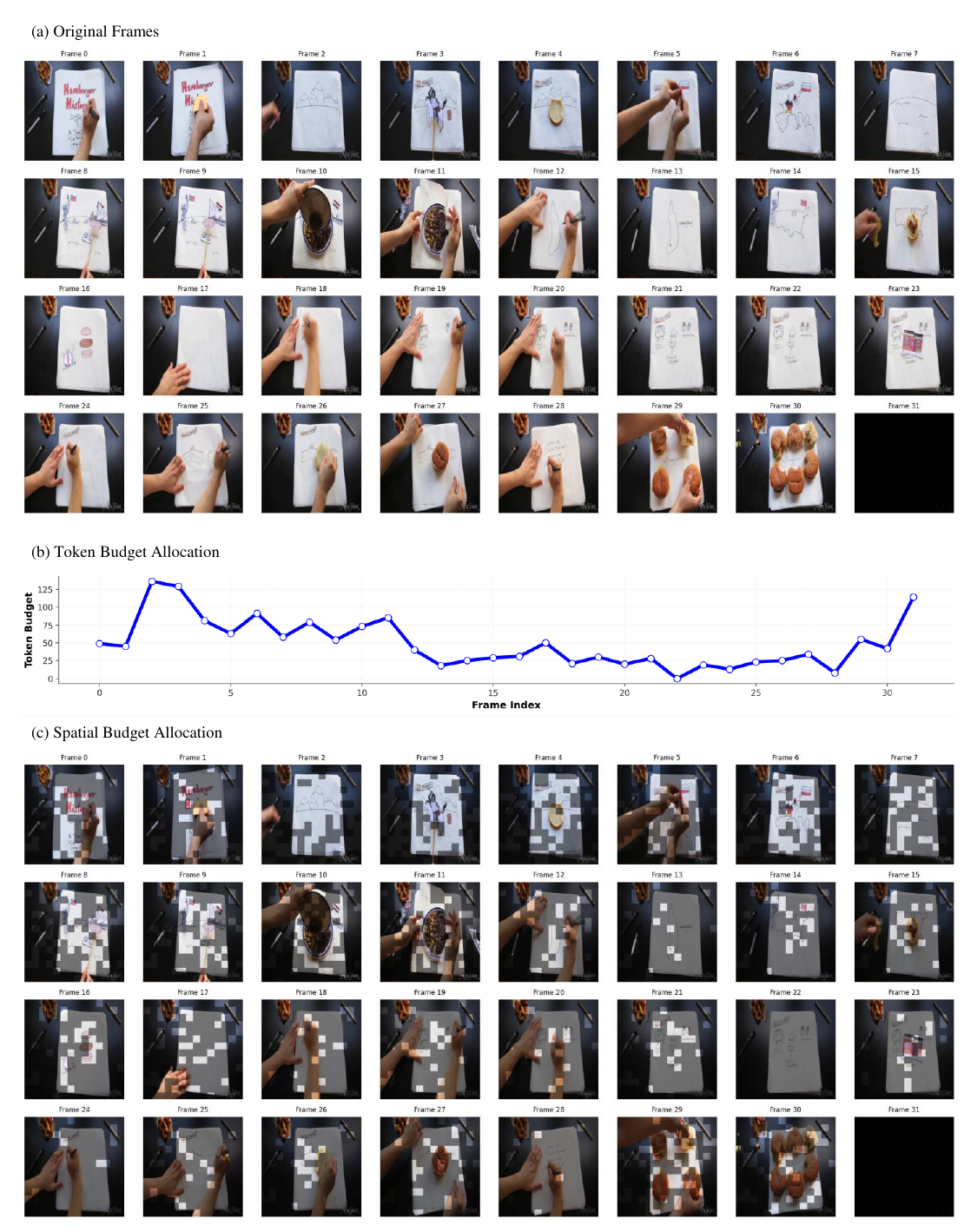}
\caption{Full 32-frame token compression visualization corresponding to the example in
Fig. 5.}
\label{fig:supple_figures_drawing}
\end{figure*}

For clarity, we provide the full 32-frame token compression visualizations corresponding to the
qualitative examples shown in Fig.~5 of the main paper. Due to space constraints, only a subset of frames was shown in the main
manuscript. The complete sequences are provided here in
Figs.~\ref{fig:supple_figures_human} and \ref{fig:supple_figures_drawing}.

\end{document}